\documentclass{ifacconf}

\usepackage{graphicx}
\usepackage[round]{natbib}

\usepackage{diagbox}
\usepackage{url}
\usepackage{amsmath}
\usepackage{amssymb}
\usepackage{float}
\usepackage[absolute]{textpos}

\makeatletter
\def\endfigure{\end@float}
\def\endtable{\end@float}
\makeatother
\let\ifacconfcaptionwidth\captionwidth
\usepackage[caption=false]{subfig}
\let\captionwidth\ifacconfcaptionwidth

\TPMargin{5pt}

\newcommand{\copyrightstatement}{
    \begin{textblock}{0.8}(0.1,0.08)    % tweak here: {box width}(leftposition, rightposition)
         \noindent
         \centering
         \copyright2023 the authors. This work has been accepted to IFAC for publication \\ under a Creative Commons Licence CC-BY-NC-ND.
    \end{textblock}
}
\begin{document}
\copyrightstatement

\begin{frontmatter}

\title{Stochastic MPC Based Attacks on Object Tracking in Autonomous Driving Systems\thanksref{footnoteinfo}}

\thanks[footnoteinfo]{This material is based upon work supported by the National Science Foundation (NSF) under Grant Number CNS-1801611.}

\author{Sourav Sinha and Mazen Farhood}

\address{Crofton Department of Aerospace and Ocean Engineering, Virginia Tech, Blacksburg, VA 24061 USA (\{srvsinha, farhood\}@vt.edu)}
%\address[Second]{Virginia Tech, Blacksburg, VA 24060 USA (email: farhood@vt.edu)}

\begin{abstract}
% Abstract of not more than 250 words.
Decision making in advanced driver assistance systems involves in general the estimated trajectories of the surrounding objects. Multiple object tracking refers to the process of estimating in real time these trajectories, leveraging for this purpose sensors to detect the objects.
This paper deals with devising attacks on object tracking in automated vehicles. The vehicle is assumed to have a detection-based object tracking system that relies on multiple sensors and uses an estimator such as a Kalman filter for sensor fusion and state estimation.
The attack goal is to  modify the object's state estimated by the victim vehicle to put the vehicle in an unsafe situation. This goal is achieved by judiciously perturbing some or all of the sensor outputs corresponding to the object of interest over a desired horizon.
A stochastic model predictive control (SMPC) problem is formulated to compute the sequence of perturbations, whereby hard constraints on the perturbations and probabilistic chance constraints on the object's state are imposed. The chance constraints ensure that some desired conditions for a successful attack are satisfied with a prespecified probability.
Reasonable assumptions are then made to obtain a computationally tractable linear SMPC program.
The approach is demonstrated on an adaptive cruise control system in a simulation environment, where successful sequential attacks are generated, leading the victim vehicle into dangerous driving situations including collisions.
\end{abstract}


\end{frontmatter}
%===============================================================================

\section{Introduction}

Detection and tracking of moving objects are key requirements for the safety-critical decision-making process in autonomous driving systems. An automated vehicle leverages onboard sensors such as cameras, LiDARs, and radars for the detection of the surrounding objects. These objects are then tracked using multiple object tracking (MOT) techniques, which construct the evolving trajectories of the detected objects.
%Multiple object tracking (MOT) techniques are then used to construct the evolving trajectories of the detected objects. Sequential adversarial attacks on object detection over some time horizon could mislead an automated vehicle with incorrect estimates of the object trajectories \citep{Ma2021sequential}, thus compromising any advanced driver assistance systems used, such as forward collision warning (FCW) and adaptive cruise control (ACC) systems, that make decisions based on these compromised trajectories. Successful attacks could lead to the victim vehicle entering dangerous and potentially grave driving situations. 
Sequential adversarial attacks on object detection over a given time horizon could mislead an automated vehicle, resulting in incorrect estimates of object trajectories \citep{Ma2021sequential}. This could compromise the effectiveness of advanced driver assistance systems, including forward collision warning (FCW) and adaptive cruise control (ACC) systems, that make decisions based on the estimated trajectories of surrounding objects. Successful attacks could lead to the victim vehicle entering dangerous and potentially grave driving situations.

In recent years, there has been a significant amount of research focused on adversarial attacks in object detection. Deep Neural Networks (DNNs) have become the state-of-the-art in computer vision, but they have been shown to be vulnerable to adversarial perturbations in visual input \citep{szegedy2013intriguing}. These perturbed inputs, often referred to as adversarial examples, can cause DNNs to generate incorrect predictions in many visual recognition tasks such as image classification, semantic segmentation, and object detection \citep{moosavi2016deepfool,eykholt2018robust,xie2017adversarial,song2018physical}.
The vulnerability of LiDAR-based perception in real-world vehicles has been extensively explored in the literature. Researchers have demonstrated the ability to fool LiDAR detectors by generating adversarial point clouds \citep{cao2019adversarial}, adding perturbations to the existing LiDAR points \citep{xiang2019generating}, or completely hiding a target vehicle from the LiDAR detector \citep{tu2020physically}. Furthermore, the radar sensors used in automated vehicles are also susceptible to adversarial attacks that can cause them to output incorrect object position and velocity estimates \citep{yan2016can, sun2021control}. Recent work has also demonstrated simultaneous attacks on both camera and LiDAR sensors \citep{cao2021invisible}.

While object detection attacks have received significant attention over the years, there has been limited research on sequential detection attacks that target the MOT system to modify/hijack the object trajectories estimated by automated vehicles. Recently, in \cite{jia2020fooling}, it was shown that the attacks that blindly target object detection alone are insufficient. MOT involves complex tasks such as data association, multi-sensor fusion, and optimal state estimation, which present additional challenges to existing object detection attacks in terms of their attack impact and detectability.

%%%%%%%%%%%%%%%%%%%%%%%%%%%%%%%%%%%%%%%%%%%%%%%%%%%%
\subsection{Related Works}

%\subsubsection{Adversarial Attacks on Object Tracking}
In a recent study \citep{jia2020fooling}, the authors investigated adversarial machine learning attacks against detection-based MOT systems in autonomous driving. The study assumed that the victim vehicle used a single camera for detecting objects in conjunction with a Kalman filter (KF). The attacks targeted consecutive camera-captured frames to deviate the object trajectory until the true object was no longer associated with the modified trajectory. These attacks, however, were limited to a small time horizon and focused solely on deviating the object trajectory, without considering the potential impact of such a modification on the autonomous driving components.

The work of \cite{Ma2021sequential} builds on the work of \cite{jia2020fooling} and explores a more realistic scenario involving a victim vehicle that employs a radar in addition to a camera. In this work, the adversary perturbs the sensor measurements to modify an object's trajectory and ultimately compromise the output of the FCW system. It utilizes model predictive control (MPC) to determine the optimal sequence of perturbations. The authors, however, make strong assumptions regarding the adversary's capabilities and do not consider the data association problem in their MPC formulation. As a result, the perturbed trajectory resulting from this attack may not be associated with the intended object, possibly compromising the effectiveness~of~the~attack. %Further research is necessary to explore more sophisticated attack strategies that consider the data association problem and the limits of the adversary's capability.

\subsection{Contributions}
This study proposes a stochastic MPC (SMPC) approach for generating effective sequential attacks on KF-based object tracking in autonomous driving systems. These attacks are in the form of bounded measurement perturbations. The proposed approach extends the MPC-based attack developed in \cite{Ma2021sequential} to the stochastic setting, with weaker and more realistic assumptions. Specifically, compared to the aforementioned MPC-based approach, this work includes the following novel additions:
\begin{enumerate}
    \item The grey-box attack setting is adopted, where the adversary has limited knowledge of the victim.
    \item The measurement perturbations are corrupted with random noise rather than being applied exactly.
    \item The adversary does not have access to the victim's sensor outputs or its estimation of an object's state and error covariance.
    \item A stochastic process model is used to better predict the benign measurements over the prediction horizon. 
    \item The temporal consistency problem is addressed to ensure appropriate data association and decrease the likelihood of an attack being detected.
\end{enumerate}
Finally, the proposed approach is demonstrated on an ACC system, where different scenarios are considered.

%%%%%%%%%%%%%%%%%%%%%%%%%%%%%%%%%%%%%%%%%%%%%%%%%%%%%%%%%%%
%The rest of the paper is organized as follows: Section {2} gives the notation and some needed background; Section {3} describes the threat model; Section {4} presents the formulation of the SMPC-based attack approach; Section {5} presents an example involving the ACC system of an AV; Section {6} gives some  concluding remarks.

\section{Preliminaries}

\subsection{Notation}
The sets of real vectors of $n$ elements and real matrices of size $n\times m$ are denoted by $\mathbb{R}^n$ and $\mathbb{R}^{n \times m}$, respectively. 
%The identity matrix of size $m \times m$ is denoted by $I_m$. 
%
The block diagonal augmentation of matrices $A_1,\, \dots,\, A_n$ is expressed as $\text{diag}(A_1,\, \dots,\, A_n)$.
Given $u\in\mathbb{R}^n$, its $1$-norm is defined as $\lVert u \rVert_{1} = \sum_{i=1}^{n}\lvert u_i \rvert$.
$\mathbb{P}(E)$ denotes the probability of an event $E$. The symbol $\mathbf{1}$ denotes a vector whose components are all one with dimension determined from  context. A multivariate Gaussian distribution with mean $\mu$ and covariance $\Sigma$ is denoted by $\mathcal{N}(\mu,\Sigma)$. $\mathbb{E}[x]$ denotes the expected value of a random variable $x$.

\begin{figure}[t]
\begin{center}
\centerline{\includegraphics[width=0.36\textwidth]{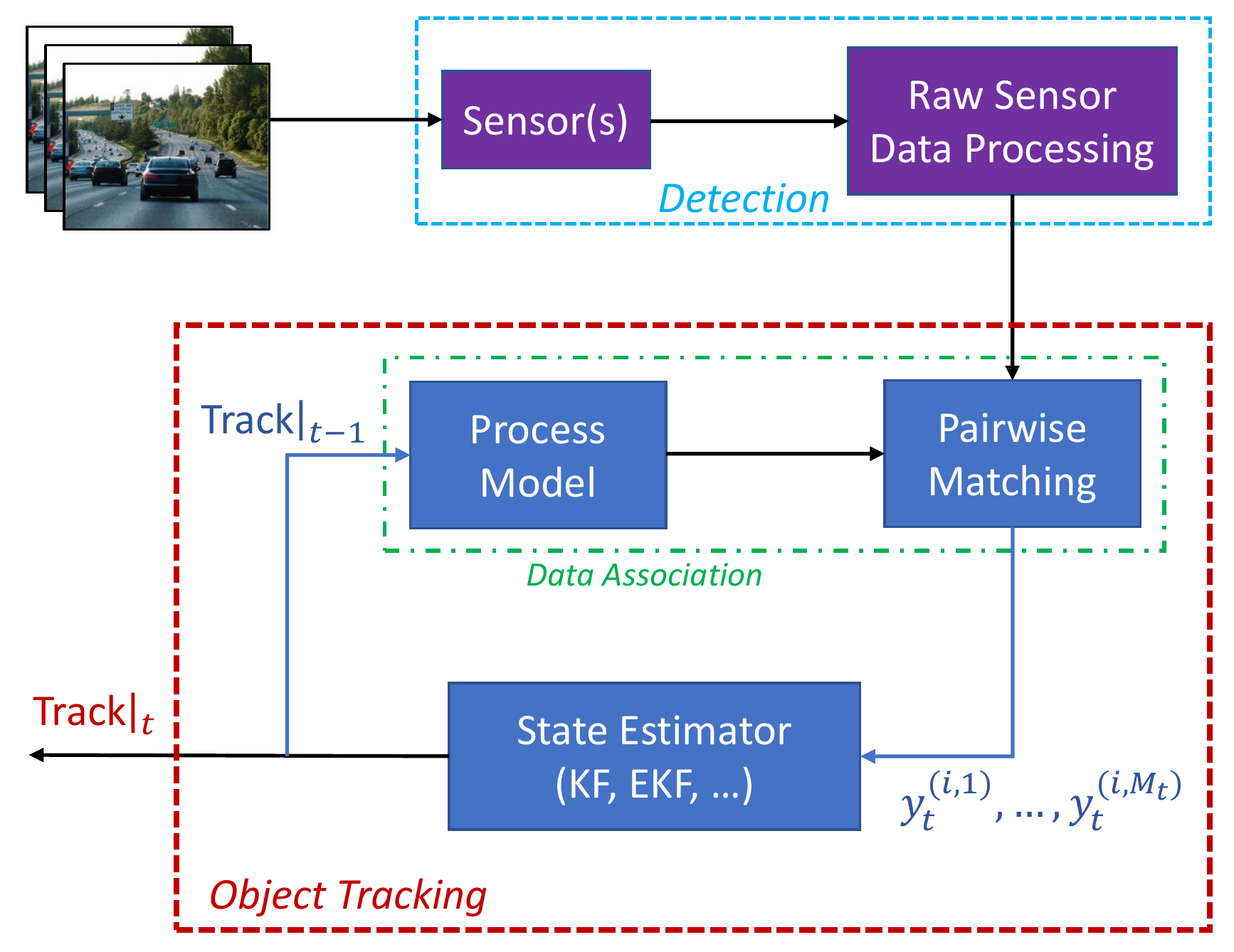}}
\caption{A procedure flow of the detection-based MOT.}
\label{figMOT}
\end{center}
\end{figure}

%%%%%%%%%%%%%%%%%%%%%%%%%%%%%%%%%%%%%%%%%%%%%%%%%%%%%%%%%
\subsection{Multi-Object Tracking} \label{MOT}

Multi-object tracking aims to locate multiple objects in a scene, identify them, and build and maintain their trajectories, called tracks \citep{luo2021multiple}. In detection-based methods, MOT leverages sensors to localize the objects which are then tracked over time to build their trajectories. The tracking process involves associating the detected objects with existing object tracks and estimating the internal state of each object from noisy sensor measurements \citep{cho2014multi}. 
%
%The data association problem in detection-based tracking is formulated as a pairwise matching problem, e.g., a Hungarian assignment problem \citep{kuhn1955hungarian} where a similarity metric measures the cost of assigning a detection to an existing track. Once the detections are assigned to the existing tracks, a state estimator such as a Kalman filter (KF) or an extended KF (EKF) is used for multi-sensor fusion and state estimation.
In detection-based tracking, the data association problem is typically framed as a pairwise matching problem, which involves assigning each detection to an existing track using a similarity metric that quantifies the cost of each assignment. The Hungarian algorithm \citep{kuhn1955hungarian} is a popular optimization method used for solving the data association problem. Once the detections are assigned to the existing tracks, a state estimator such as a KF or an extended KF (EKF) is employed for multi-sensor fusion and state estimation. 
%
%These estimators use a motion model to describe the relative dynamic behavior of the object and a measurement model to relate the object's state to the sensor output. 

Figure~\ref{figMOT} provides a pictorial representation of the tracking algorithm. At any given time $t$, the sensors generate a set of detections $\{ O_t^{(i,1)},\, O_t^{(i,2)},\, \dots,\, O_t^{(i,n_{i,t})} \}$. Here, $O_t^{(i,j)}$ represents the $j^{th}$ detection made by the $i^{th}$ sensor at time $t$.
%
%Track$|_{t-1}$ consists of the tracks of all the objects estimated till the time instant $t-1$. The process (motion and measurement) model predicts the sensor observations corresponding to those objects/tracks at time $t$. The distance between the predicted measurement corresponding to an existing track and an actual detection is used as the cost of assigning that detection to the existing track. A pairwise matching problem is then solved for each sensor $i$ to assign detections $O_t^{(i,j)}$ for $j=1,\dots,n_{i,t}$ to the existing tracks in Track$|_{t-1}$.  
Track$|_{t-1}$ is a collection of estimated tracks of all objects of interest up until time instant $t-1$. Using the process (motion and measurement) model, sensor observations for the existing tracks at time $t$ are predicted. The cost of assigning a detection to an existing track is determined by calculating the distance between the predicted measurement corresponding to the track and the actual detection. For each sensor $i$, a pairwise matching problem is then solved to assign the detections $O_t^{(i,j)}$ to the existing tracks in Track$|_{t-1}$.
Let $y_t^{(i,j)}$ denote the $i^{th}$ sensor detection that is assigned to the $j^{th}$ track in Track$|_{t-1}$, where $j\in\{1,\,2,\,\dots,\,M_{t-1}\}$ and $M_{t-1}$ is the total number of object tracks at time $t-1$. If no detection corresponding to the $i^{th}$ sensor is assigned to the $j^{th}$ track, then $y_t^{(i,j)}$ will be empty. If no detection is assigned to an existing track continuously for a pre-specified number of time steps, the track is deleted, indicating that the corresponding object has left the scene.
For $j=M_{t-1}+1,\,\dots,\,M_{t}$, $y_t^{(i,j)}$ represents the detections corresponding to the new objects that have entered the scene at time $t$.
By using measurements from all the sensors at the current step that  correspond to an object, the state estimator calculates the best estimate of that object's current state, denoted by $\bar{x}_{t}$, and updates the track to get Track$|_{t}$. This process is repeated at each time step to construct the evolving trajectories of the surrounding objects.

\section{Threat Model}

The objective of an adversary is to manipulate the state of an object that is being estimated by a victim vehicle over a certain period of time. By doing so, the adversary aims to force the victim vehicle into dangerous and potentially grave driving situations. This goal is achieved by judiciously perturbing some or all of the sensor measurements corresponding to the object of interest over the desired horizon. In this work, we are not concerned with devising the mechanism to perturb the sensor measurements; instead, we focus on developing an effective approach to compute the optimal sequence of the bounded perturbations. To this end, we make some assumptions necessary to develop an SMPC-based attack approach. 

\emph{Assumption 1:} 
In the scenarios of interest, the object and the victim vehicle dynamics can be approximated satisfactorily by linear time-invariant (LTI) models over a small attack horizon. 

Although the system dynamics are typically nonlinear, in certain scenarios, it is possible to satisfactorily approximate the system dynamics using linear models. For instance, a stochastic constant acceleration model can reasonably approximate the dynamics of a vehicle driving straight on a highway.

\emph{Assumption 2:} 
The victim vehicle uses a KF estimator for multi-sensor fusion and state estimation.

The KF is a commonly used estimator. For an LTI model with Gaussian process and measurement noise, a KF gives the optimal minimum-mean-squared-error estimate of the system’s state. However, when dealing with a nonlinear model, an EKF is typically used. In the scenarios of interest, an adversary can  use a KF and obtain satisfactory results since a linear model is an appropriate approximation of the nonlinear system in these scenarios and an EKF reduces to a linear KF when the model is linear.

\emph{Assumption 3:}
The KF parameters, namely, the process noise covariance and measurement noise covariance, are known to the adversary.
%The justification for this assumption is that the attacks will target standard vehicles, which an adversary can acquire to reverse-engineer these KF parameters.

The attacks are expected to target standard vehicles. An adversary can acquire such vehicles and perform reverse engineering to extract  the values of the KF parameters.

% \subsubsection{Assumption 4}
% The environment is sparse in the sense that the chances of the perturbed detection of one object associating with the track of some other object are low.\\
% A sparse environment is assumed to ensure that the data association is not affected by the attack, i.e., the perturbed detection of the object of interest will not associate with the track of a different object and vice versa. It is also possible to make the attack work in a non-sparse environment if the pairwise matching algorithm for data association is considered in the problem formulation.

\emph{Assumption 4:}
%The adversary has similar sensors to the victim to measure the position and velocity of the object of interest relative to the victim vehicle.\\
%Unlike the previous work in \cite{Ma2021sequential}, the victim’s sensor outputs and its estimation of the state and error covariance of the object are not accessible to the adversary. We also do not assume that the adversarial perturbation can be applied exactly without any error.
The adversary is equipped with sensors similar to those of the victim to measure the position and velocity of the object of interest relative to the victim.

Unlike the previous work in \cite{Ma2021sequential}, the adversary does not have access to the victim's sensor outputs or its estimation of the state and error covariance of surrounding objects. Furthermore, it is not assumed that the adversarial perturbation can be precisely applied without~any~error.

% \subsubsection{Assumption 6}
% The measurement perturbation error can be modeled as white Gaussian noise.\\
% This assumption is not limiting, as the approach can be adjusted to incorporate perturbation errors that are characterized differently.

%%%%%% Comment

%The pairwise similarity cost considered in the data association problem is proportional to the spatial discrepancy between the detection $y_k$ and the process model prediction $\left(h(f(\bar{x}_{k-1}))\right)$. The error between the detection and the process model prediction is called the KF innovation or residual. The residual is, sometimes, also used by the defenders to detect attacks and weed out temporally inconsistent sensor measurements. Even though we do not impose any constraints on the residual, we minimize a weighted residual in the optimization problem so that the attacks go undetected and the data association remains unaffected by the attack. Assumption 6 ensures that perturbed detection of an object does not associate with the trajectory of a different object and vice versa.

\section{Approach}

Assume without loss of generality that the attack starts at $k=0$ and ends at $k=N$. The adversary needs to determine the optimal perturbation sequence $(\delta_0,\,\delta_{1},\,\dots,\,\delta_N)$ that, if applied to the output $(y_0,\,y_{1},\,\dots,\,y_N)$ of the victim's sensors, will result in an incorrect object trajectory $(\bar{x}_0,\,\bar{x}_{1},\,\dots,\,\bar{x}_N)$ estimated by the victim vehicle that satisfies the adversary's desired constraints while minimizing an objective function. 
To compute the optimal sequence of perturbations, we formulate an SMPC problem. At each time step $k$, we solve a chance-constrained optimization problem over a fixed prediction horizon of $M$ steps in order to determine the perturbation sequence $(\delta_{k|k},\,\delta_{k+1|k},\,\dots,\,\delta_{k+M-1|k})$. Here, the symbol $z_{j|i}$ denotes the prediction of the variable $z$ at time step $j$ using the information available up to time step $i$. The first perturbation from the sequence is applied at time step $k$, i.e., $\delta_k = \delta_{k|k}$, and at the next time step, the optimization problem is reformulated and solved again over a shifted horizon; see \cite{mesbah2016} for an overview of SMPC.
%see \cite{mesbah2016, MPCbook2017} for an overview on MPC and SMPC.

To approximate the response of the victim's state estimator to an adversarial perturbation sequence, the adversary uses a KF with an LTI process model given as
\begin{subequations}
\label{eqnProcessModel} 
\begin{align}
x_{k+1} = Ax_k + w_k \label{eqnProcessModela}  \\ 
y_k = Cx_k + v_k \label{eqnProcessModelb} 
\end{align}
\end{subequations}
where $x_k \in \mathbb{R}^{n_x}$ is the object's state relative to the victim  and $y_k \in \mathbb{R}^{ n_y} $ is the concatenation of measurements from all the victim's sensors at time $k$. 
The process and measurement Gaussian white noise sequences are represented by $w_k \sim \mathcal{N}(0,Q)$ and $v_k \sim \mathcal{N}(0,R)$, respectively. Without loss of generality, the adversary assumes that the KF is in a steady state, which simplifies the state estimation recursion process to the following linear equation:
\begin{equation}
    \label{eqnKFEst}
    \bar{x}_k = \bar{A}\bar{x}_{k-1} + Ky_k
\end{equation}
where $\bar{A}=A-KCA$ and $K$ is the steady-state Kalman gain given by $K = PC^T(CPC^T + R)^{-1}$.
Here, $P$ is the solution of the following algebraic Riccati equation:
\begin{equation*}
    \label{eqnRiccati}
    P = APA^T - APC^T(CPC^T + R)^{-1}CPA^T + Q.
\end{equation*}

%%%%%%%%%%%%%%%%%%%%%%%%%%%%%%%%%%%%%%%%%%%%%%%%%%%%%%%%%%%%%%%%%%%%%%%%%%%%%%
\subsection{Measurement Prediction}
The victim's sensor outputs over the prediction horizon are required to predict the state estimated by the victim corresponding to an adversarial perturbation sequence. At each time step, the adversary obtains a noisy measurement from its own sensors, with mean $y^a_k$ and covariance $R^a$, and uses this value as an uncertain approximation of the victim's current measurement. The adversary then predicts the victim's future measurements using the stochastic process model \eqref{eqnProcessModel}. Since the process model is LTI, the predicted measurements can be directly expressed as a function of the benign initial state $\hat{x}_k$ using a linear recursion:
\begin{equation*}
      \hat{y}_{k+i|k} =
     C\left(A^i\hat{x}_k + A^{i-1}w_k  + \dots + A^{0}w_{k+i-1} \right) + v_{k+i}.
\end{equation*}
A linear combination of Gaussian random vectors is Gaussian; thus, we can rewrite the predicted measurements as
\begin{equation}
    \label{eqnypred}
      \hat{y}_{k+i|k} =
  \begin{cases}
     y_k^a + \gamma_0 & \text{if $i=0$} \\
     CA^i\hat{x}_k + \gamma_i  & \text{if $i = 1,2,\dots,M-1$}.
  \end{cases}
\end{equation}
Here, $\gamma_i$ denotes a Gaussian white noise sequence, where $\gamma_i \sim \mathcal{N}(0,\Psi_i)$ with $\Psi_0 = R^a$ and, for $i=1,2,\dots,M-1$,
\begin{equation*}
    \label{eqncovy}
    \Psi_{_i} = C\left(\sum_{j=0}^{i-1}A^jQ(A^j)^T\right)C^T + R.
\end{equation*}
%%%%%%
\textit{Benign State Approximation:} The adversary approximates the benign state of an object by processing the measurements obtained from its sensors. The adversary can use denoising filters and derivative techniques such as total-variation regularization \citep{chartrand2011numerical} to obtain a good approximation $\hat{x}_k$ of the benign state $x_k$ from the set of noisy measurement data $(y_{k-L}^a,\,y_{k-L+1}^a,\,\dots,\,y_{k}^a)$.

%%%% 
\textit{Perturbed Measurement:} The adversarially perturbed measurement over the prediction horizon is given as
\begin{equation}
   \label{eqnypert}
    \tilde{y}_{k+i|k} = \hat{y}_{k+i|k} + W_{s}(\delta_{k+i|k} + \epsilon_{k+i|k})
\end{equation}
%where $\delta_{k+i|k}$ is the adversary-desired perturbation and $\epsilon \sim \mathcal{N}(0,\Xi)$ is the perturbation error. $W_{s} = \text{diag}(a_1,\,\dots,\,a_{n_y})$ where $a_i$ is $1$ if the $i^{th}$ element of $y$ can be perturbed by the adversary or 0 otherwise. 
where $\delta$ is the applied perturbation, $\epsilon$ is the random perturbation error, and $W_s$ is a coefficient matrix.

%%%%%%%%%%%%%%%%%%%%%%%%%%%%%%%%%%%%%%%%%%%%%%%%%%%%%%%%%%%%%%%%%%%%%%%%%%%%%
\subsection{Predicting the Adversarial State Estimate}
The adversary uses the state estimation process \eqref{eqnKFEst} and the adversarially perturbed measurement obtained using \eqref{eqnypert} to predict an object's state estimated by the victim vehicle. By implementing a linear recursion, the predicted state estimate over the prediction horizon $(i=0,1,\dots,M-1)$ can be defined~as
\begin{equation}
    \label{eqnxpred}
    \bar{x}_{k+i|k} = \bar{A}^{i+1}\bar{x}_{k-1|k} + \bar{B}_{i+1} \begin{bmatrix} \tilde{y}_{k|k}^T & \dots & \tilde{y}_{k+i|k}^T \end{bmatrix}^T
\end{equation}
where $\bar{B}_{i+1} = \begin{bmatrix} \bar{A}^{i}K & \dots & \bar{A}^{0}K \end{bmatrix}$ and $\bar{x}_{k-1|k} = \bar{x}_{k-1}$ is the adversary's approximated output of the victim's state estimator at time step $k-1$. This output can be computed by the adversary using the following relation:
\begin{equation}
    \label{eqnInitStateEst}
    \bar{x}_{k-1} = \bar{A}^k\bar{x}_{-1} + \bar{B}_k \begin{bmatrix} \Tilde{y}_{0}^T & \dots & \Tilde{y}_{k-1}^T \end{bmatrix}^T
\end{equation}
where $\bar{x}_{-1}$ is the benign state estimated by the victim vehicle before the attack starts and $\Tilde{y}_{j} = \Tilde{y}_{j|j}$ for $j=0,1,\dots,k-1$ 
are the uncertain approximations of the actual perturbed measurements seen by the victim vehicle. The benign state estimate $\bar{x}_{-1}$, which is not known to the adversary, is an estimate of the true state ${x}_{-1}$, and $\hat{x}_{-1}$ computed by the adversary is also a good approximation of the true state; hence, we can write
\begin{equation}
    \label{eqnInitStateEst0}
    \bar{x}_{-1} = \hat{x}_{-1} + \eta
\end{equation}
where $\eta \in \Lambda$ is a bounded uncertainty that accounts for the discrepancy between the two estimates.
Using \eqref{eqnInitStateEst} and \eqref{eqnInitStateEst0}, we can rewrite \eqref{eqnxpred} as
\begin{equation}
\label{eqnxpred2}
\begin{split}
    \bar{x}_{k+i|k} &= 
    \bar{A}^{k+i+1}\left(\hat{x}_{-1} + \eta \right) \\
    &+ 
     \bar{B}_{k+i+1} \begin{bmatrix} \Tilde{y}_{0|0}^T & \dots &  \Tilde{y}_{k|k}^T & \Tilde{y}_{k+1|k}^T & \dots & \Tilde{y}_{k+i|k}^T\end{bmatrix} ^T.
     \end{split}
\end{equation}

%%%%%%%%%%%%%%%%%%%%%%%%%%%%%%%%%%%%%%%%%%%%%%%%%%%%%%%%%%%%%%%%%%%%%%%%%%%%%%%%%%%%%%%%%%%%%%%%
\subsection{Constraints}
To conform to the limitations of the adversary, we impose the following hard constraint on the perturbations:
\begin{equation}
   \label{eqnPertCon}
    -\mathbf{1}\Delta \leq W_{{p}}\delta_{k+i|k} \leq \mathbf{1}\Delta
\end{equation}
where  $\Delta$ is a prespecified positive scalar  and $W_{{p}}$ is a diagonal matrix used to normalize the perturbation. 
The adversary also imposes constraints on the predicted state estimate to ensure that some desired conditions for a successful attack are satisfied. In this work, we consider polytopic constraints, which are defined as
\begin{equation*}
    \label{eqnxconst}
    F\bar{x}_{k+i|k} \leq h, \quad i=0,1,\dots,M-1
\end{equation*}
where $F \in \mathbb{R}^{s \times n_x}$ and $h \in \mathbb{R}^{s}$ are specified by the adversary. The predicted state estimates, however, are stochastic in nature, and imposing hard constraints on them could lead to an infeasible optimization problem. To address this issue, we relax the hard constraints with soft joint chance constraints of the form
\begin{equation}
    \label{eqnxconst2}
    \mathbb{P}(F\bar{x}_{k+i|k} \leq h) \geq 1- \alpha
\end{equation}
where $\alpha$ is the risk of constraint violation.

%%%%%%%%%%%%%%%%%%%%
\subsubsection{Reformulation of the Chance Constraint:}
Using Boole's inequality, the joint chance constraint \eqref{eqnxconst2} can be replaced with $s$ individual chance constraints of the form
\begin{equation}
    \label{eqnxconst2b}
    \mathbb{P}(f_{j}^T\bar{x}_{k+i|k} \leq h_{j}) \geq 1- {\alpha}/{s}, \quad j = 1,\,\dots,\,s
\end{equation}
where $f_{j} \in \mathbb{R}^{n_x}$ is the $j^{th}$ row of $F$ and $h_{j}$ is the $j^{th}$ element of $h$. The individual chance constraints can then be replaced with the deterministic linear constraints
\begin{equation}
    \label{eqnxconst3}
    f_{j}^T \mathbb{E}\left[\bar{x}_{k+i|k}\right] \leq  h_{j} - \Omega(\alpha/s)\sqrt{f_{j}^T\Sigma_{k+i|k}f_{j}}
\end{equation}
where $\Sigma_{k+i|k}$ is the covariance of $\bar{x}_{k+i|k}$, defined in \eqref{eqnxpred2}. For an arbitrarily distributed state estimate, take  $\Omega(p) = \sqrt{(1-p)/p}$. If the predicted state estimate is Gaussian, i.e., the perturbation error has a Gaussian distribution, less conservative inequalities can be obtained by taking  $\Omega(p) = \Phi^{-1}(1-p)$, where $\Phi$ is the cumulative distribution function of the standard normal distribution; see \cite{okamoto2018,paulson2020} for details.
%
% The predicted state estimate $\bar{x}_{k+i|k}$ is Gaussian with mean $\mathbb{E}\left[\bar{x}_{k+i|k}\right]$ and covariance $\Sigma_{k+i|k}$. By decomposing the joint chance constraint using the approach discussed in \cite{okamoto2018}, we replace \eqref{eqnxconst2} with $s$ individual deterministic linear constraints:
% \begin{equation}
%     \label{eqnxconst3}
%     f_{j}^T \mathbb{E}\left[\bar{x}_{k+i|k}\right] \leq  h_{j} - \sqrt{f_{j}^T\Sigma_{k+i|k}f_{j}}\Phi^{-1}\left(1 - \frac{\alpha}{s}\right)
% \end{equation}
% where $ j = 1,\dots,s $, $f_{j} \in \mathbb{R}^{n_x}$ is the $j^{th}$ row of $F$, $h_{j}$ is the $j^{th}$ element of $h$, and $\Phi$ is the cumulative distribution function of the standard normal distribution. 
%
%
The constraint in \eqref{eqnxconst3} should be satisfied for all possible values of $\eta \in \Lambda$. Denoting $\zeta_{k+i|k} = \mathbb{E}\left[\bar{x}_{k+i|k}\right] -\bar{A}^{k+i+1}\eta  $ and 
$\Gamma^{(j)}_{k+i|k} \left(\alpha \right) = \Omega(\alpha/s)\sqrt{f_{j}^T\Sigma_{k+i|k}f_{j}} + \max_{\eta \in \Lambda} f_{j}^T\bar{A}^{k+i+1}\eta$, a sufficient condition for \eqref{eqnxconst3} is 
\begin{equation}
\label{eqnxconst4}
    f_{j}^T \zeta_{k+i|k} \leq   h_{j} - \Gamma^{(j)}_{k+i|k} \left(\alpha \right).
\end{equation}

%%%%%%%%%%%%%%%%%%%%%%%%%%%%%%%%%%%%%%%%%%%%%%%%%%%%%%%%%%%%%%%%%%%%%%%%%%%%%%%%%%%%%%%%%%%%%%%%%%%
\subsection{Objective Function}
It is possible to formulate and solve a multi-criterion optimization problem, but, in this work, we just consider a bi-criterion optimization problem with the cost function $(J_1,J_2)$, where $J_1=\sum_{i=0}^{M-1}J_1^{(i)}$ and $J_2=\sum_{i=0}^{M-1}J_2^{(i)}$,  with %defined as $J  = \sum_{i=0}^{M-1}{J_1^{(i)} + \lambda J_2^{(i)}}$, where 
\begin{subequations}
\begin{align}
    J_1^{(i)} &= {\left\lVert \mathbb{E}\left[\tilde{y}_{k+i|k} - CA\bar{x}_{k+i-1|k} \right]  \right\rVert_1} \quad \text{and} \\
    J_2^{(i)} &= \mathbb{E}\left[c^T\bar{x}_{k+i|k}\right].
    \label{eqnobj}
\end{align}
\end{subequations}
The first objective is to minimize the KF residual, which serves as a useful tool for defenders to weed out temporally inconsistent sensor measurements. Also, as outlined in Section~\ref{MOT}, the position residual is used as the cost of assigning a detection to an existing track. Therefore, minimizing $J_1$ helps to reduce the likelihood of the attack getting detected and the data association getting affected. 
It is also possible to address the pairwise matching algorithm in our framework to ensure correct data association of the object of interest. This can be done by imposing convex conditions on the pairwise matching costs; however, we do not impose these conditions to avoid the added conservatism associated with chance constraints. In environments with sparse objects, such as highways  during off-peak hours or in rural areas, the likelihood of correctly matching the object of interest can be significantly improved by minimizing $J_1$ alone. This is because the cost of matching a perturbed detection with the track of a different object and vice versa will be relatively high in a sparse environment.

The second objective, like the constraint on the state estimate, is problem-dependent and is defined by the adversary to improve the impact of the attack. Since we have two (most probably) competing objectives, the problem will not have an optimal solution \citep{boyd2004}. In this case, Pareto optimal solutions can be obtained by solving a scalarized problem with cost function $J=J_1+\lambda J_2$ for different positive values of $\lambda$. Then, trade-off analysis can be conducted to choose the best suited value of $\lambda$.

%%%%%%%%%%%%%%%%%%%%%%%%%%%%%%%%%%%%%%%%%%%%%%%%%%%%%%%%%%%%%%%%%%%%%%%%%%%%%%%%%%%%%%%%%%%%%%%%%%%
\subsection{Optimization Problem}

The finite horizon linear optimization problem that the adversary needs to solve at each time step $k$ for $k=0,1,\dots,N$ can now be expressed as
\begin{subequations}
\label{eqnopt} 
\begin{align*}
\min_{\delta_{k|k},\dots,\delta_{k+M-1|k}} \quad &  \sum_{i=0}^{M-1}{J_1^{(i)} + \lambda J_2^{(i)}} \\
 \text{subject to} \quad & \text{Equations}~\eqref{eqnypred}- \eqref{eqnInitStateEst0}  \\ 
 & \text{Constraints}~\eqref{eqnPertCon}~\text{and}~\eqref{eqnxconst4} \\
 & \forall ~ j = 1,\dots,s, \quad \forall ~ i=0,1,\dots,M-1.
\end{align*}
\end{subequations}
Note that the adversary cannot access the output of the victim's detector or state estimator. As a result, the adversary can not account for the random missed detections or ID switches in data association, which are common issues even in the absence of an adversary. These events can introduce errors in the estimation of the victim’s belief \eqref{eqnxpred2} and reduce the effectiveness of the attack. However, we believe that our approach will provide a certain degree of robustness to these discrepancies, given that the estimate $\bar{x}_{k+i|k}$ is stochastic and the attacks are limited to a small horizon. While the reduction in attack efficiency may be significant when these events occur frequently, it should be negligible when the probability of these events is low.

\section{Attack Evaluation}

We demonstrate the developed attack approach on an adaptive cruise control (ACC) system. 

\begin{table}[tb]
\begin{center}
\caption{Parameters of the ACC system.}
\label{tab:ACCparams}
\begin{tabular}{ccccccc}
\hline
$k_{sc}$ & $k_{v}$ & $k_d$ & $d_{def}$ & $T_g$ & $a_{max}$ & $a_{min}$\\
\hline
$0.5$ & $0.9$ & $1.4$ & $2~\mathrm{m}$ & $0.6~\mathrm{s}$ & $2~\mathrm{m/s^2}$ & $-6~\mathrm{m/s^2}$\\
\hline
\end{tabular}
\end{center}
%\vskip -0.3in
\end{table}

%%%%%%%%%%%%%%%%%%%%%%%%%%%%%%%%%%%%%%%%%%%%%%%%%%%%%%%%%%%%%%%%%%%%%%%%%%%%%%%%%%%%%%%%%%%%%%%%%%%%%%%%%%%%%%%%%%%%%%%%%%%%%%%%%%%%%
\subsection{Adaptive Cruise Control System}

The ACC system is an advanced driver assistance system that automatically regulates the longitudinal acceleration of a vehicle to assist the human driver. Its primary objective is to maintain a driver-set speed $v_{set}$, i.e., $v_{ego} = v_{set}$, while ensuring that a safe distance $d_{safe}$ is maintained between the lead vehicle and the ego vehicle.
\begin{figure}[H]
  \centering
  \includegraphics[width=0.4\textwidth]{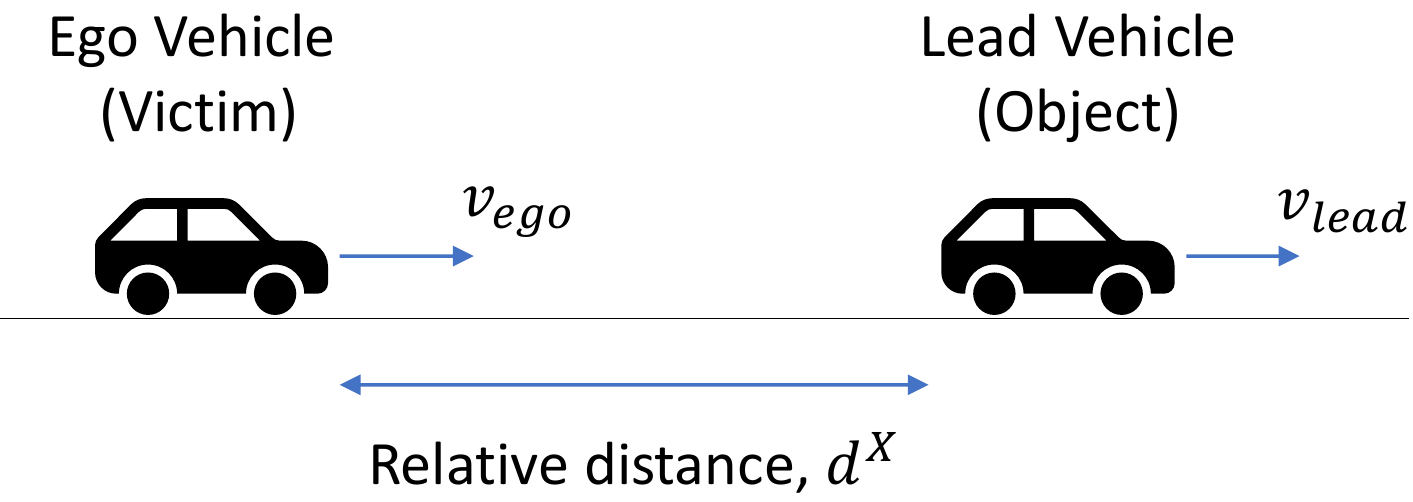}
  \caption{ACC equipped victim vehicle in highway scenario.}
  \label{figACC}
\end{figure}
For ease of demonstration of the attack, we consider the classical ACC system used in \cite{MatlabACC},
% , similar to the ones considered in \cite{koschi2019computationally,amoozadeh2015platoon,gunter2020commercially}, 
where the acceleration is computed using the following control~law:
\begin{equation}
    \label{eqnACC}
      a^X_{des}=
     \min{\left(a_v, a_d   \right)}.
\end{equation}
Here, $a_v = k_{sc}(v_{set} - v_{ego})$, $a_d = k_vv^X + k_d(d^X -d_{safe})$, $v^X$ and $d^X$ are the relative speed  and distance of the lead vehicle estimated by the ego vehicle along the driving direction, and $k_{sc}$, $k_v$, and $k_d$ are the control gains.
The safe distance is defined as $d_{safe} = d_{def} + T_{g}v_{ego}$, where $d_{def}$ is the default gap and $T_g$ is the time gap.  The time dependence of the variables in the preceding is suppressed for simplicity.
We assume that there is no reaction delay to provide a best case scenario for the ACC system. Some parameters are modified to make our setup more consistent with real-world scenarios. We change $d_{def}$ from $5\, \mathrm{m}$ to $2\, \mathrm{m}$ and $T_g$ from $1.5\, \mathrm{s}$ to $0.6\, \mathrm{s}$
to get a reasonable safe gap of $20\, \mathrm{m}$ ($>4 \times$ average car length) instead of the conservative $50\, \mathrm{m}$ gap at $30\, \mathrm{m/s}$ driving speed. Inspired by other papers and to make the attack more challenging, we strengthen the victim's deceleration capabilities by changing $a_{min}$ from $-3\,\mathrm{m/s^2}$  to $-6\,\mathrm{m/s^2}$. The control gains are then fine-tuned through simulations to improve the performance of the victim's ACC system; see Table~\ref{tab:ACCparams} for the values.

%%%%%%%%%%%%%%%%%%%%%%%%%%%%%%%%%%%%%%%%%%%%%%%%%%%\
%%%%%%%%%%%%%%%%%%%%%%%%%%%%%%%%%%%%%%%%%%

\begin{figure*}[tb]
\centering
\subfloat[\label{fig:1a}]{\includegraphics[width=0.33\textwidth]{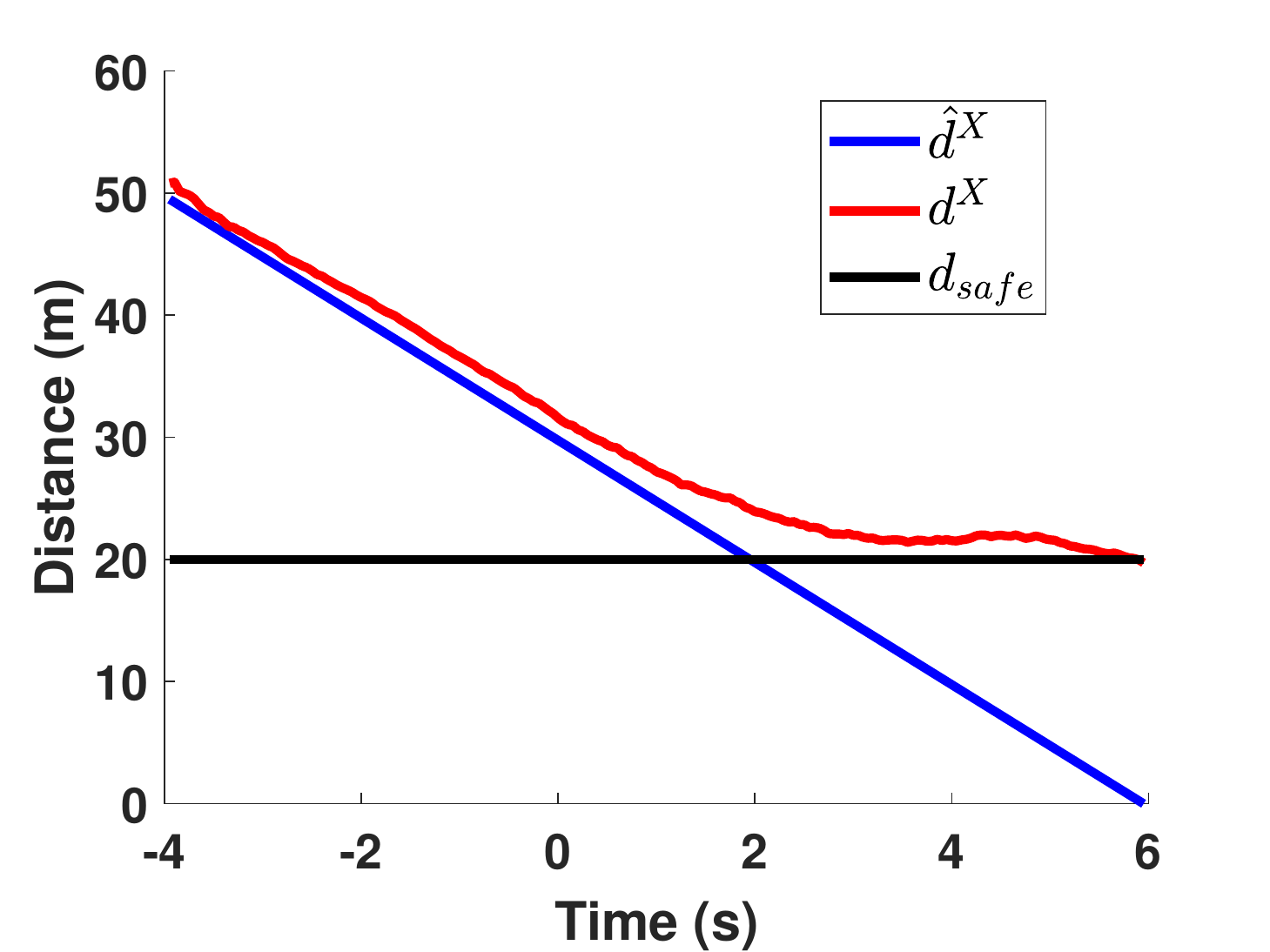}}\hfill
\subfloat[\label{fig:1b}] {\includegraphics[width=0.33\textwidth]{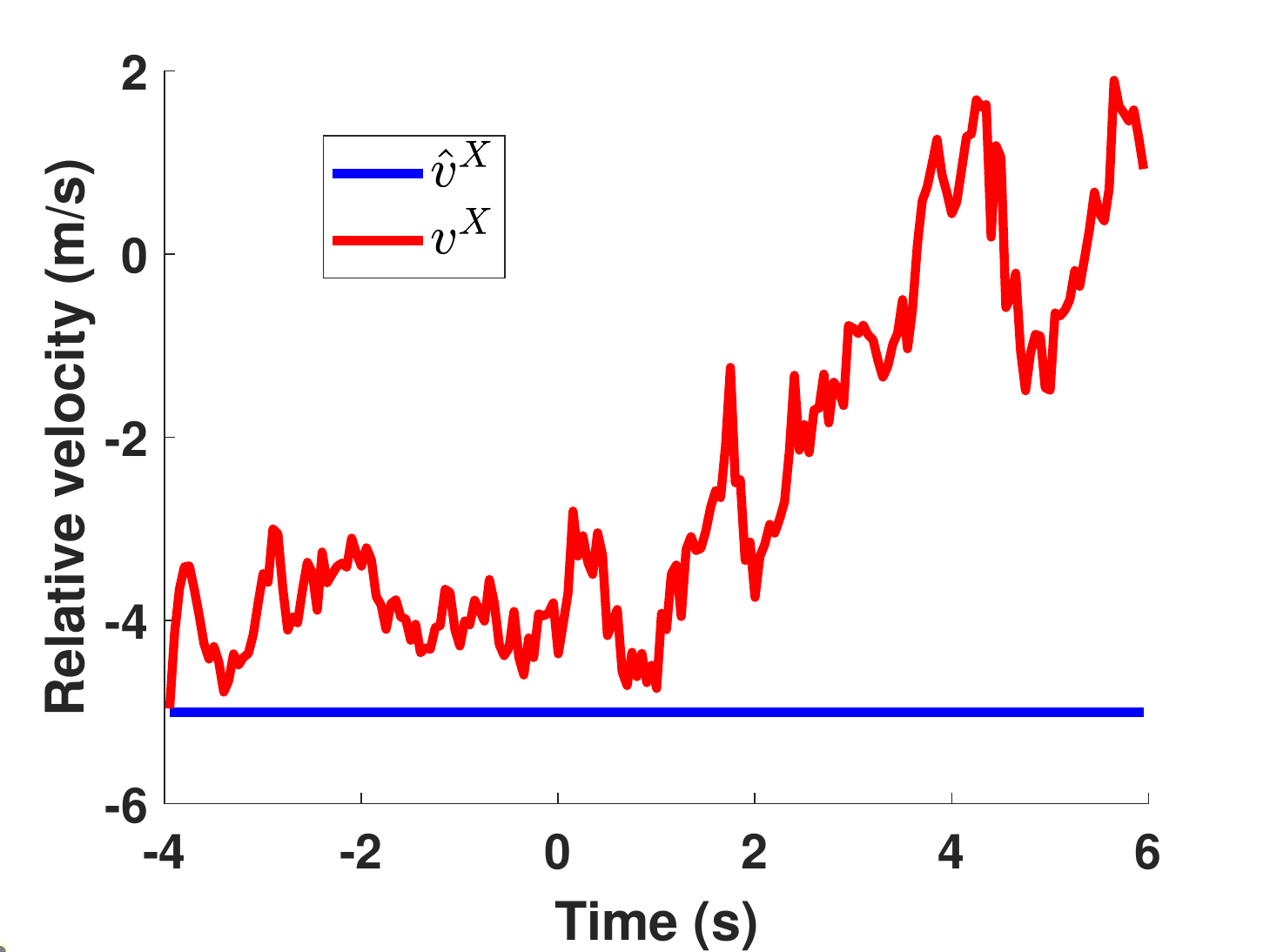}}\hfill
\subfloat[\label{fig:1c}]{\includegraphics[width=0.33\textwidth]{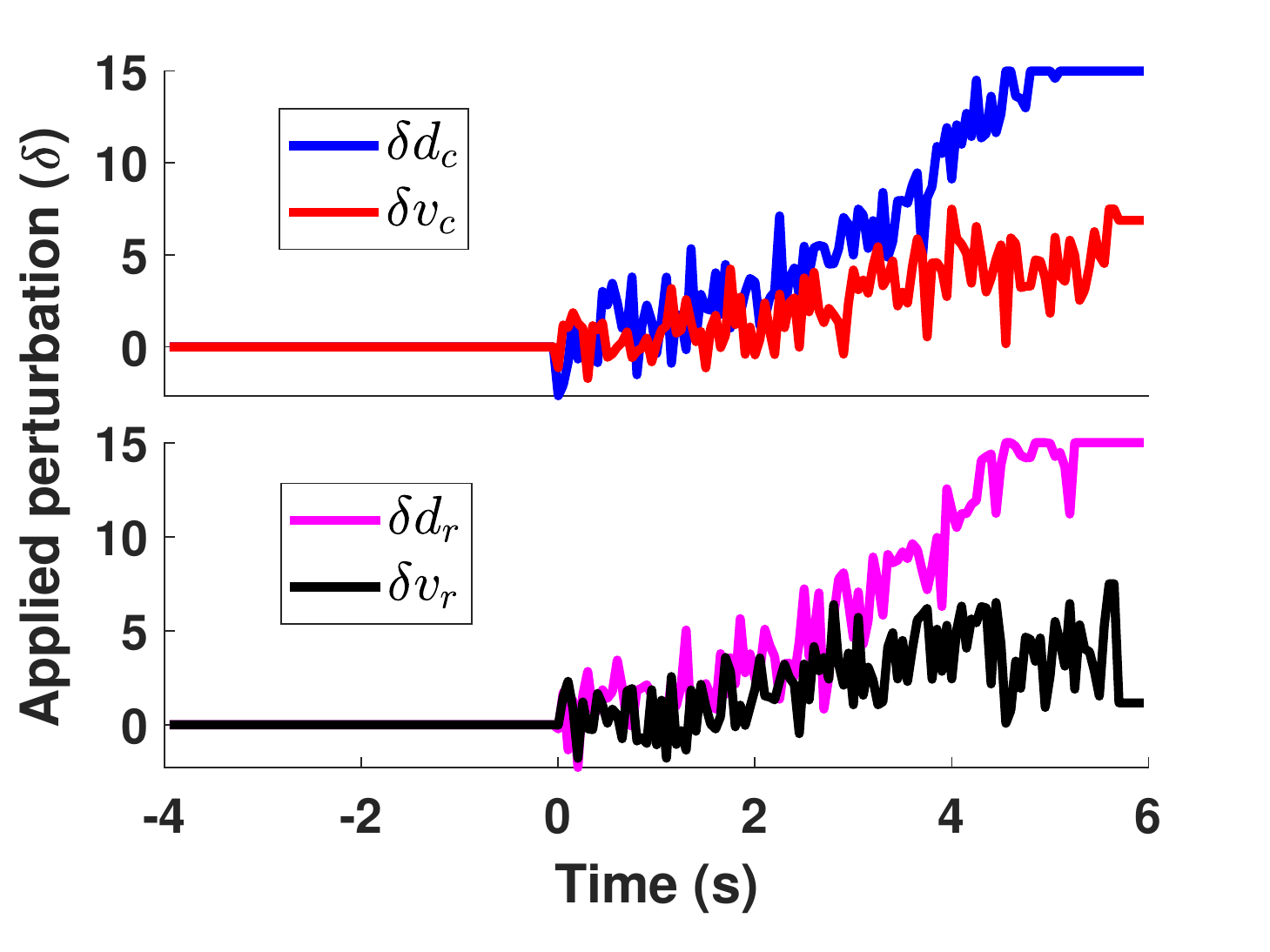}}
\caption{Results of the attack based on the first strategy with $\Delta=15$ and $W_s = \text{diag}(1,1,1,1).$ } 
\label{fig:1}
\end{figure*}

%%%%%%%%%%%%%%%%%%%%%%%%%%%%%%%%%%%%%%%%%%%%%%%%%%%%%%%%%%%%%%%%%%%%%%%%%%%%%%%%%%%%%%%%%%%%%%%%%%%%%%%%%%%%%%%%%%%%%%%%%%%%%%%%%%%%%
\subsection{Simulation Environment}
We consider the same victim vehicle as the one in \cite{MatlabFCW,Ma2021sequential}, which uses a camera and a radar to measure the distances and relative velocities of the surrounding objects along ($X$-axis) and perpendicular ($Y$-axis) to the driving direction.
%Both these sensors measure the distances and relative velocities of the surrounding objects along ($X$-axis) and perpendicular ($Y$-axis) to the driving direction. The victim uses a stochastic constant acceleration process model with a KF estimator (see Appendix~\ref{CAM}). 
We consider a situation in which the lead vehicle (object) is traveling at a speed of $25\,\mathrm{m/s}$ and the ego vehicle (victim) has the ACC system engaged and is traveling at the driver-set speed of $30\,\mathrm{m/s}$. We assume that both vehicles are driving straight on a highway and there are no other vehicles in close proximity to the lead vehicle. 
We simulate the victim and the lead vehicles in MATLAB. The focus of this paper is on determining the sequence of perturbations that achieves some desired response when applied to the victim’s noisy measurements. This problem is one part of the attack mechanism, which encompasses other equally important and challenging parts. With enough uncertainties incorporated into our simulation environment to make it as realistic as possible for our purposes, we believe that the MATLAB simulations are sufficient for evaluating the computed perturbation sequences.

For the data association problem, the victim uses the Euclidean norm of the KF position residual as the similarity metric. 
%Since we only consider a single object in the scene, we do not solve a pairwise matching problem. 
A detection is assigned to an existing track if the cost of the detection assignment, i.e., the similarity metric, is less than a threshold value of $10\,\mathrm{m}$. If no detection is  assigned to a track, the process model is used to estimate the state. The track is deleted (object move-out) if no detection is assigned to it continuously for $1$ second. Similarly, detections  not assigned to any tracks are considered as false positives. A new track (new object move-in) is  generated if unassigned detections have been reported at least 80\% of the times over a $0.5\,\mathrm{s}$ interval. 

The adversary approximates the unknown process model used by the victim vehicle with a stochastic constant acceleration model. The adversary is equipped with the same sensors as the victim vehicle, which are used to measure the lead vehicle's position and velocity relative to the victim vehicle. Since the ACC system relies solely on the longitudinal dynamics and the longitudinal and the lateral dynamics in the constant acceleration model are decoupled, the adversary only focuses on the longitudinal mode of the constant acceleration model, where $x_k = [d_k^{X},\, v_k^{X},\, a_k^{X}]^T \in \mathbb{R}^3$  and $y_k = [d_k^{X,c},\, v_k^{X,c},\, d_k^{X,r},\, v_k^{X,r}]^T \in \mathbb{R}^4$ are the state and the sensor output evaluated at time $k$, respectively; see Appendix~\ref{CAM} for details. 
We take the covariance of the perturbation uncertainty to be $ \text{diag}(0.2,\,0.2,\,0.2,\,0.2)$. The risk of constraint violation is chosen as $\alpha =0.05$. 
The uncertainty $\eta$ is bounded as follows: $-\bar{\eta}\preceq \eta\preceq \bar{\eta}$, where $\bar{\eta} = [1.0,\,0.5,\,0.5]^T$ and $\preceq$ denotes the componentwise inequality. The normalization matrix for the perturbation is chosen as $W_p = \text{diag}(1,\,2,\,1,\,2)$. A prediction horizon of $10$ steps is selected for the SMPC problem. Through measurement perturbation attacks, the goal of the adversary is to compromise the ACC system such that the distance of the object falls below the safe distance. To quantify the impact/efficiency of the attack, we~use~the~metric
\begin{equation*}
    \xi = \max \left(0, \,\, \max_{0\leq k\leq N} 1-\frac{\hat{d}^X_k}{{d}_{safe}\bigr|_k} \right)
\end{equation*}
where $\hat{d}^X$ is the true distance of the object from the victim and $N+1$ the length of the attack horizon. A $\xi$-value of one implies the maximum impact (collision), and a value of zero implies that the attack did not have any impact.

%%%%%%%%%%%%%%%%%%%%%%%%%%%%%%%%%%%%%%%%%%%%%%%%%%%%%%%%%%%%%%%%%%%%%%%%%%%%%%%%%%%%%%%%%%%%%%%%%%%%%%%%%%%%%%%%%%%%%%%%%%%%%%%%%%%%%
\subsection{Attack Strategies and Results}

One strategy to attack the ACC system is to modify the object's state estimated by the victim such that the victim keeps traveling at the constant speed of $v_{set}$, i.e, $a^X_{des}\bigr|_{v_{ego}=v_{set}}=0$, throughout the attack. This can be ensured by imposing the chance constraints \eqref{eqnxconst2} with $F = -\begin{bmatrix} k_d & k_v & 0 \end{bmatrix}$ and $h=-k_d(d_{def}+T_gv_{set})$. It is assumed that the ACC system parameters are known to the adversary. In this strategy, the second objective function \eqref{eqnobj} is not required.
In the simulations, we start the attack after the true distance of the object becomes $30~\mathrm{m}$, i.e., $\hat{d}^X_{-1}=30$. We take $W_s = \text{diag}(1,1,1,1)$ to perturb all the sensor measurements. If the attack works perfectly, the victim vehicle should collide with the object in $6~\mathrm{s}$. It is important to note that for small $\Delta$ values, it may not be possible to ensure a zero acceleration for $6~\mathrm{s}$. In such cases, the constrained optimization problem will become infeasible after some time. We carry out this attack for different values of $\Delta$, and the attack is stopped when the problem becomes infeasible or when a collision occurs, whichever comes first. 
We end the simulation once the attack is stopped and  compute the attack efficiency $(\xi)$ for each test case; see Table~\ref{tab:Att1}. From the table, we observe that a minimum delta of $15$ is required for a collision. For this attack case, the variations of the object's true distance $(\hat{d}^X)$ and the distance estimated by the victim $({d^X})$ are shown in Fig.~\ref{fig:1a}. The variations of the true $(\hat{v}^X)$ and victim's estimated $({v^X})$ relative velocities and the applied perturbation sequences are shown in Fig.~\ref{fig:1b} and Fig.~\ref{fig:1c}.

\begin{table}[t]
\caption{Variation of the efficiency ($\xi$) of the first attack strategy with $\Delta$.}
  \label{tab:Att1}
\begin{center}
\begin{tabular}{ccccc}
\hline
$\Delta$  & $6$ & $9$ & $12$ & $15$ \\
\hline
$\xi$  & $0.37$ & $0.65$ & $0.73$ & $1.00$ \\
\hline
\end{tabular}
\end{center}
\end{table}

%%%%%%%%%%%%%%%%%%%%%%%%%%%%%
\begin{table}[t]
\caption{Variation of the efficiency $(\xi)$ of the second attack strategy with $\lambda$ and $\Delta$ when all the measurements are perturbed.}
\label{tbl:Att2}
\begin{center}
\begin{tabular}{lccccc}
\hline
\diagbox{$\lambda$}{$\Delta$} & $3$  & $6$ & $9$ & $12$ & $15$\\
\hline
$0.5$ & 0.36 & 0.60  & 0.87  & 0.97  & 1.00   \\
$1$ & 0.47  & 0.77 & 1.00   &1.00  &1.00   \\
$2$ & 0.46 &  0.78 & 1.00   &1.00  &1.00   \\
$5$ & 0.48 & 0.81 & 1.00   &1.00  & TD   \\
$10$ &0.48  &  0.82 & 1.00   &1.00  & TD   \\
\hline\end{tabular}
\end{center}
\end{table}

%%%%%%%%%%%%%%%%%%%%%%%%%%%%%%%%%%%%%%%%%%

\begin{figure*}[tb]
\centering
\vskip -0.15in
\subfloat[\label{fig:2a}]{\includegraphics[width=0.33\textwidth]{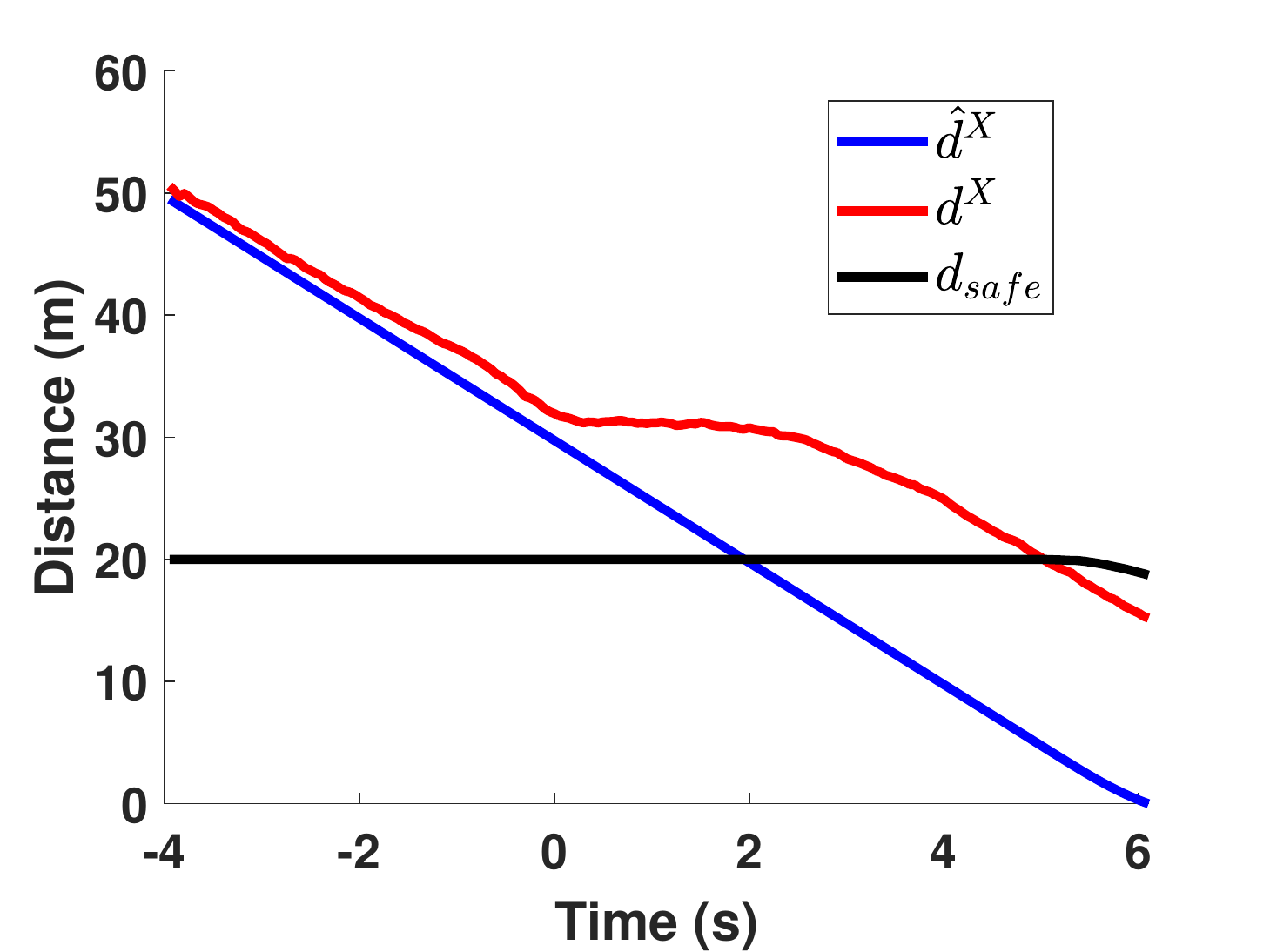}}\hfill
\subfloat[\label{fig:2b}]{\includegraphics[width=0.33\textwidth]{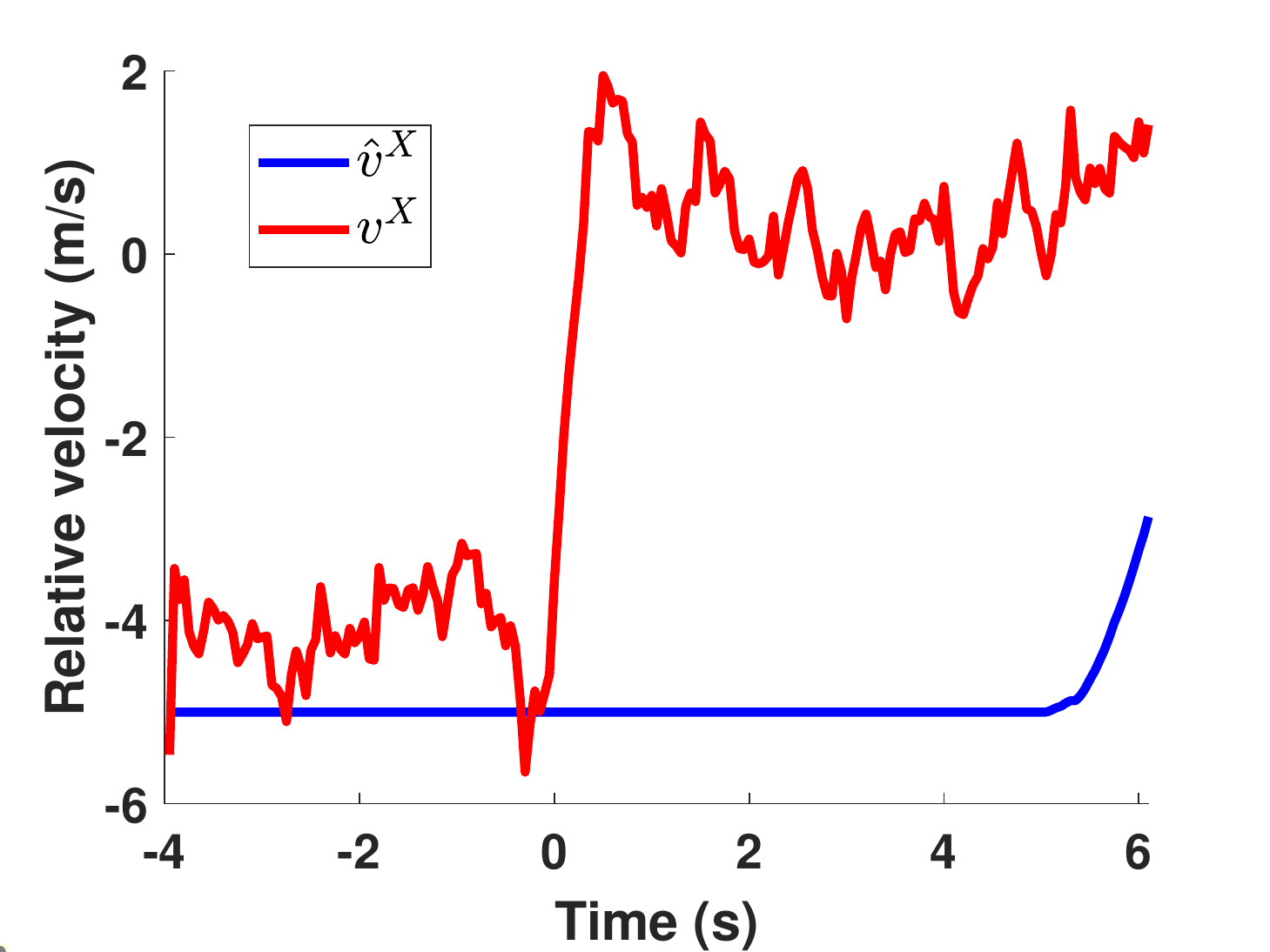}}\hfill
\subfloat[\label{fig:2c}]{\includegraphics[width=0.33\textwidth]{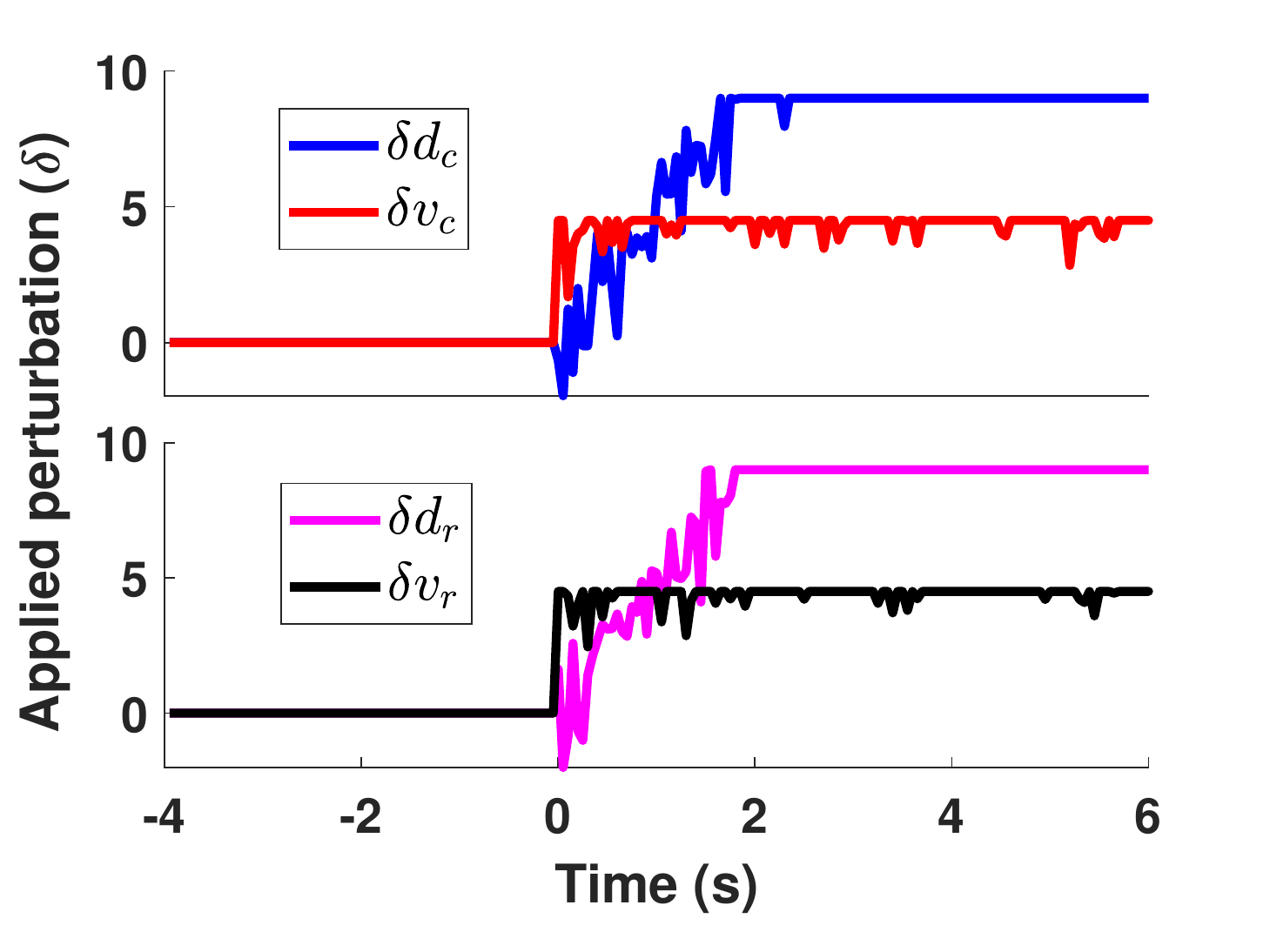}}
\caption{Results of the attack based on the second strategy with $\Delta=9$, $\lambda=1$, and $W_s = \text{diag}(1,1,1,1)$.} \label{fig:2}
\end{figure*}

%%%%%%%%%%%%%%%%%%%%%%%%%%%%%%%%%%%%%%%%%%
\begin{figure*}[tb]
\centering
\subfloat[\label{fig:3a}]{\includegraphics[width=0.33\textwidth]{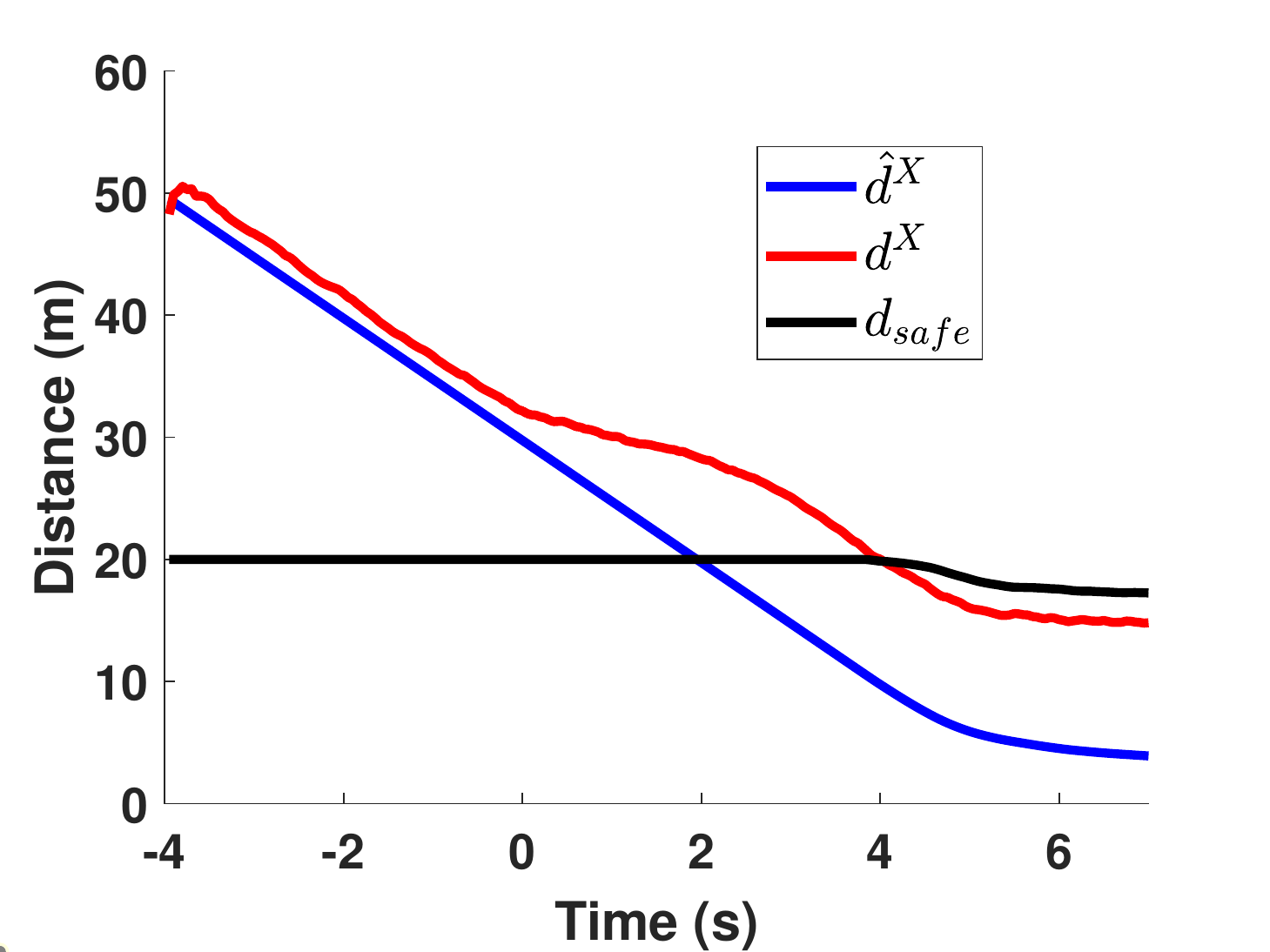}}\hfill
\subfloat[\label{fig:3b}] {\includegraphics[width=0.33\textwidth]{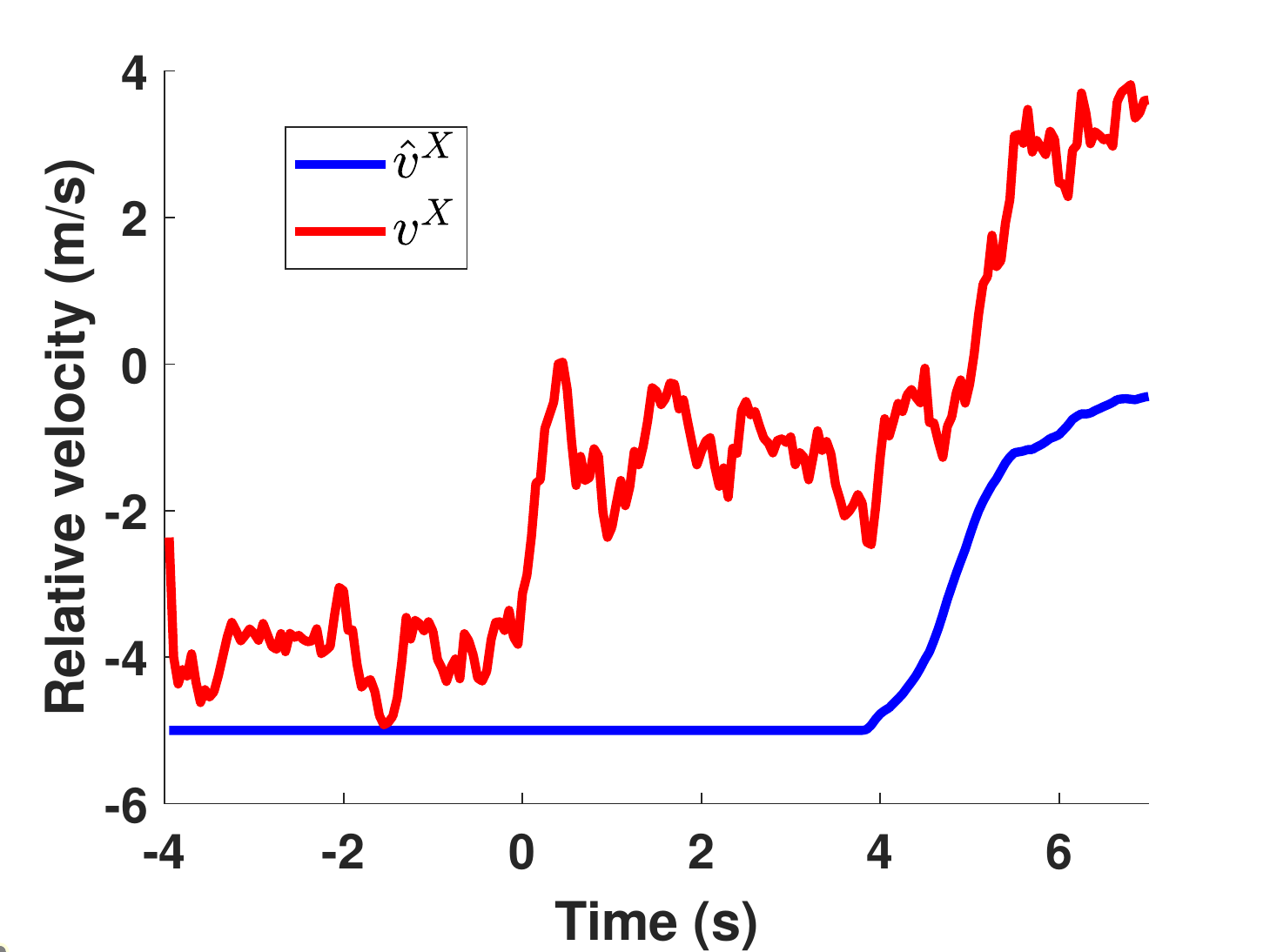}}\hfill
\subfloat[\label{fig:3c}]{\includegraphics[width=0.33\textwidth]{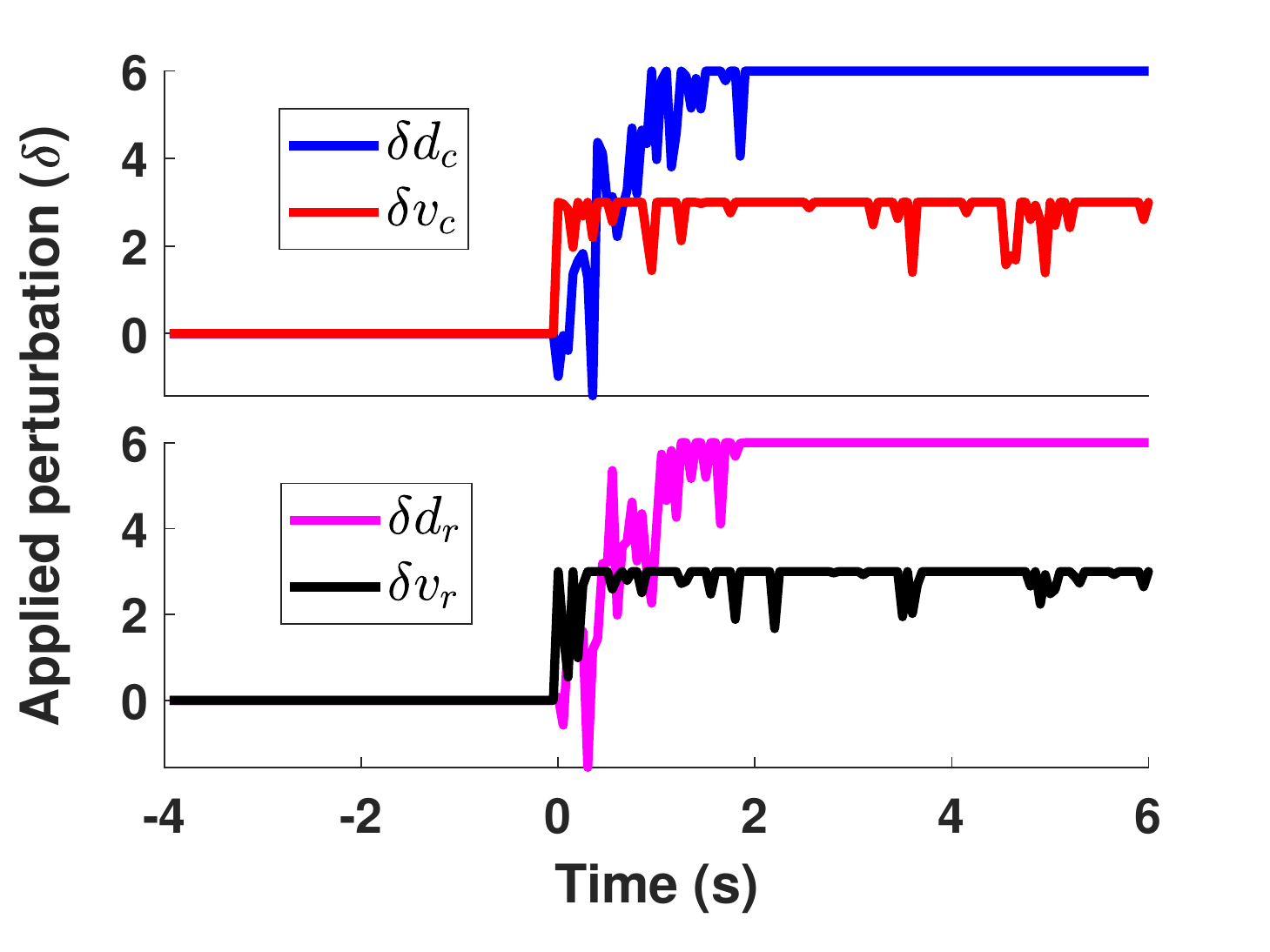}}
\caption{Results of the attack based on the second strategy with $\Delta=6$, $\lambda=1$, and $W_s = \text{diag}(1,1,1,1)$.} \label{fig:3}
\end{figure*}

The aforementioned attack strategy has some limitations. The attack stops when the optimization problem becomes infeasible, i.e., when a zero acceleration is not possible. However, it may still be possible to ensure that the deceleration of the victim is smaller than what it should have been if there was no attack. This goal could be achieved by appropriately shaping $h$ over the attack horizon. A simpler strategy is to maximize the victim vehicle's acceleration during the attack instead of constraining it. To do this, we take $c = -\begin{bmatrix} k_d & k_v & 0 \end{bmatrix}^T$ in \eqref{eqnobj}. As there are no chance constraints, the SMPC problem reduces to an MPC problem with a cost function expressed as the positive weighted sum of two objective functions. We carry out multiple attacks for different values of $\Delta$, and for each $\Delta$, we obtain multiple Pareto optimal solutions by varying $\lambda$. We take $d^X_{-1}=30$ and $W_s = \text{diag}(1,1,1,1)$. Since the problem is always feasible, we stop the attack after $7~\mathrm{s}$ if a collision does not occur. The values of the attack efficiency $\xi$ for different combinations of $\Delta$ and $\lambda$ are presented in Table~\ref{tbl:Att2}. Large values of $\lambda$ are favorable to increase the efficiency of the attack; however, when $\Delta$ is large, more weight on the KF residual, i.e., a smaller $\lambda$, is required to ensure that the perturbed detection is associated with the modified track. As seen in the table, the modified track gets deleted (TD) when $\Delta=15$ and $\lambda \geq 5$. The variations of the distances, relatives velocities, and perturbation sequences for $\Delta=9$ and $\Delta=6$ are shown in Fig.~\ref{fig:2} and Fig.~\ref{fig:3}, respectively. With the second attack strategy, a collision is possible with $\Delta=9$ (see Fig.~\ref{fig:2a}). Even though the deceleration is not zero throughout the attack, the adversary ensures that it is small enough (see Fig.~\ref{fig:2b}) to lead to a collision.

%%%%%%%%%%%%%%%%%%%%%%%%
\begin{table}[tb]
\begin{center}
\caption{Variation of the efficiency $(\xi)$ of the second attack strategy with $\Delta$ and the subsets of the measurements being perturbed.}
\label{tbl:Att2b}
\begin{tabular}{lccccr}
\hline 
\diagbox{$W_s$}{$\Delta$}  & $6$ & $9$ & $12$ & $15$\\
\hline
%$\text{diag}(1,1,1,1)$ & 0.77 & 1.00   &1.00  &1.00    \\
$\text{diag}(1,1,0,0)$ & 0.32 & 0.32 & 0.34  & 0.34   \\
$\text{diag}(1,0,1,0)$ & 0.44 & 0.44  & 0.52  & 0.56   \\
$\text{diag}(0,1,0,1)$ & 0.43 & 0.58  & 0.72  & 0.96   \\
\hline 
\end{tabular}
\end{center}
\end{table}

To study the effect of $W_s$, we fix $\lambda = 1$ and consider three values of $W_s$: (1) $W_s = \text{diag}(1,1,0,0)$, i.e., all camera measurements are perturbed; (2) $W_s = \text{diag}(1,0,1,0)$, i.e., the distance measurements from both sensors are perturbed; and (3) $W_s = \text{diag}(0,1,0,1)$, i.e., the velocity measurements from both  sensors are perturbed. The attack efficiencies for these test cases are given in Table~\ref{tbl:Att2b}. As expected, the efficiency decreases when not all measurements are being perturbed. From Table~\ref{tbl:Att2b}, we see that the attack is most impactful when the velocities are perturbed, and the attack is least impactful when only one sensor is attacked. The $\lambda$ values can be increased for a better attack efficiency; however, high $\lambda$ values can also create issues with the data association problem. These outcomes and observations  depend on the measurement noise covariance of each sensor and may change if the covariances are different. From an adversary's point of view, attacking sensors with smaller covariance, i.e., higher confidence, is more favorable.

%%% comment

%For the approximation of the benign state at each time step, a \textit{weak adversary} uses a set of true measurement data of cardinality $20$. To denoise the measurements, a $6^{th}$-order lowpass digital Butterworth filter with a normalized cutoff frequency of $0.3$ is used. The position and velocity states are obtained by taking the mean of denoised measurements from both sensors and the acceleration is obtained by differentiating the velocity data using the total variation regularized derivative \cite{brunton2016discovering} method. The simulation is carried out in MATLAB and the optimization problem is solved using the YALMIP toolbox \cite{lofberg2004yalmip} with MOSEK \cite{aps2019mosek} as the solver. 

\section{Conclusion}

This work provides an SMPC-based approach for generating sequential  attacks in the form of bounded measurement perturbations to compromise KF-based object tracking in autonomous driving systems. An SMPC problem is formulated to compute the optimal sequence of perturbations, whereby hard constraints on the perturbations and probabilistic chance constraints on the object’s state are imposed. The chance constraints are then reformulated to obtain a linear SMPC program. The proposed  approach is demonstrated on an ACC system of an automated vehicle.

The focus in this work has been on scenarios where the nonlinear process model can be approximated satisfactorily by a linear one. Future work will consider more general scenarios in which a linear approximation is not appropriate. In these cases, linear parameter-varying  models  \citep{toth2010modeling} that constitute a good approximation of the nonlinear dynamics over some envelope will be used.
%, and robust control tools along with MPC-based methods will be investigated. 
We will test our approach in high-fidelity virtual simulation environments to evaluate its efficacy under a range of conditions. In addition, we will analyze its performance on victim vehicles equipped with attack detection algorithms to assess its effectiveness in real-world scenarios.

%==============================================

\bibliography{ifacconf}
\appendix

\section{Constant Acceleration Model} \label{CAM}

A constant acceleration model is a linear process model that can approximate reasonably well the dynamics of a vehicle driving straight on a highway. In this model, the state of the object is represented as 
\[
x_k = [d_k^{X},\, v_k^{X},\, a_k^{X}, \, d_k^{Y},\, v_k^{Y}, \, a_k^{Y}]^T \in \mathbb{R}^6
\]
where $d$, $v$, and $a$ denote the object's relative distance, velocity, and acceleration, respectively, with $X$ and $Y$ designating the longitudinal and the lateral driving directions. The automated vehicle considered for attack demonstration uses a camera and a radar for detecting objects. Both of these sensors measure the relative distance and velocity of objects in the longitudinal and lateral directions, i.e., 
\[
y_k = [d_k^{X,c},\, v_k^{X,c},\, d_k^{Y,c},\, v_k^{Y,c}, \, d_k^{X,r},\, v_k^{X,r},\, d_k^{Y,r},\, v_k^{Y,r}]^T \in \mathbb{R}^8
\]
where the superscripts $c$ and $r$ are for camera and radar, respectively. The state and the output matrices for this model are defined as $A = \text{diag}(\bar{A},\bar{A})$ and $C = \begin{bmatrix}\bar{C}^T & \bar{C}^T \end{bmatrix}^T $, respectively, where
\begin{equation*}
    \bar{A} = 
    \begin{bmatrix} 1 & dt & dt^2/2 \\
                    0 & 1  & dt \\
                    0 & 0 & 1\end{bmatrix} ~~ \text{and}~~
    \bar{C} = 
    \begin{bmatrix} 1 & 0 & 0 & 0 & 0 & 0 \\
                    0 & 1 & 0 & 0 & 0 & 0 \\
                    0 & 0 & 0 & 1 & 0 & 0 \\
                    0 & 0 & 0 & 0 & 1 & 0 \end{bmatrix}.
\end{equation*}
Here, $dt$ is the sampling time and is chosen to be $0.05\,\mathrm{s}$ (i.e., $20$ measurements per second). The process and measurement noise covariances used by the victim vehicle are \[Q = \text{diag}(\bar{Q},\bar{Q}) \ \mbox{and} \ R = \text{diag}(2,2,2,100,2,2,2,100)\]
respectively, where 
\begin{equation*}
    \bar{Q}  = 
    \begin{bmatrix} dt^4/4 & dt^3/2 & dt^2/2 \\
                    dt^3/2 & dt^2 & dt \\
                    dt^2/2 & dt & 1\end{bmatrix}.
\end{equation*}
%

%
%It is interesting to note that the longitudinal and the lateral dynamics in the constant acceleration model are decoupled. For the longitudinal only process model with $x_k = [d_k^{X},\, v_k^{X},\, a_k^{X}]^T \in \mathbb{R}^3$ as the state and $y_k = [d_k^{X,c},\, v_k^{X,c},\, d_k^{X,r},\, v_k^{X,r}]^T \in \mathbb{R}^4$ as the sensor output, the state and the output matrices and the noise covariances are defined as
%
%\begin{equation*}
%    A = 
%    \bar{A}, ~~
%    C = 
%    \begin{bmatrix} 1 & 0 & 0  \\
%                    0 & 1 & 0  \\
%                    1 & 0 & 0  \\
%                    0 & 1 & 0  \end{bmatrix}, ~~ Q = \bar{Q}, ~~\text{and}~~ R = \text{diag}(2,2,2,2).
%\end{equation*}
%

\end{document}